\newcommand{\templateVersion}{v0.9.7}
\documentclass[a4paper,11pt]{article}
\usepackage[T1]{fontenc}
\usepackage[utf8]{inputenc}
\usepackage{lmodern}
\usepackage{graphicx}
\usepackage[hidelinks]{hyperref}
\usepackage{csquotes}
\usepackage{array}
\usepackage{tabularx}
\usepackage{booktabs}
\usepackage{multirow}
\usepackage{graphicx} %rotatebox et al
\usepackage{amsmath} %equations

\newcommand{\articleTitle}{TESSA: A Governance-Oriented Custom GPT for SME Guidance under the EU AI Act: Design, Implementation, and Evaluation}
\newcommand{\articleSubtitle}{\templateVersion}%typically not needed
\newcommand{\articleAuthors}{
    Chalisa Veesommai Sillberg\(^{a,}\)\authorFootnoteOrcid{chalisa.sillberg@tuni.fi}{0000-0001-8400-7469},
    Kai-Kristian Kemell\(^{a,}\)\authorFootnoteOrcid{KaiKristian.Kemell@tuni.fi}{0000-0002-0225-4560}
    Pekka Sillberg\(^{a,}\)\authorFootnoteOrcid{pekka.sillberg@tuni.fi}{0000-0003-2573-4775},
    Katri Harjuveteläinen\(^{b,}\)\authorFootnoteOrcid{katri.harjuvetelainen@protonmail.com}{0009-0009-2056-1562},
     Mika Saari\(^{a,}\)\authorFootnoteOrcid{mika.saari@tuni.fi}{0000-0001-7677-2355},
    Pekka Abrahamsson\(^{a,}\)\authorFootnoteOrcid{pekka.abrahamsson@tuni.fi}{0000-0002-4360-2226}
}
\newcommand{\articleAffiliation}{
    \(^{a}\) Tampere University, Faculty of Information Technology and Communication Sciences, Pori, Finland \\
    \(^{b}\) Boutique Legal Consulting, Helsinki, Finland
}
\newcommand{\articleDate}{October 22, 2026}

\newcommand{\articleNotes}{Notes: \\
    \begin{itemize}
        \item This is the author's version of the work
        \item This is a pre-print version of an article accepted to the 17th International Conference on Software Business (ICSOB 2026)
        \item The final version accepted to publishing the Springer Lecture Notes in
Business Information Processing (LNBIP)
    \end{itemize}
}
\newcommand*{\authorFootnote}[1]{\footnote{email: \url{#1}}}
\newcommand*{\authorFootnoteOrcid}[2]{\footnote{email: \url{#1}; ORCID iD: \href{https://orcid.org/#2}{[#2]}}}
\usepackage{xspace}

\usepackage[
    backend=biber,bibencoding=utf8, %instead of bibtex
    language=auto,%
    style=numeric,%
    uniquename=init,% modifies the printing of cited entries
    uniquelist=false,% modifies the printing of cited entries
    maxnames=2,% modifies the printing of cited entries
    minnames=1,% modifies the printing of cited entries
    bibstyle=ieee,%
    maxbibnames=10, % default: 3, et al.
    natbib=true]{biblatex}
\begin{document}

% This is overall setup file that controls the page numbering and format of title page and main matter (article).
% Typically it is not needed to modify anything here

% BibLaTex stuff
%\renewcommand*{\bibname}{References} % has to be declared here after preamble

%*******************************************************
% Frontmatter
%*******************************************************
%%% The title pages
\pagenumbering{roman}\setcounter{page}{1} % set roman pagenumbering for other front matter
% Including the cover page about ArXiv/preprint/etc.
%*******************************************************
% Titlepage
%******************************************************
\begin{titlepage}

%******************************
% HEADER FIGURES AND LOGOS ETC.
\begin{figure}[ht]
    \centering
    \includegraphics[height=2em]{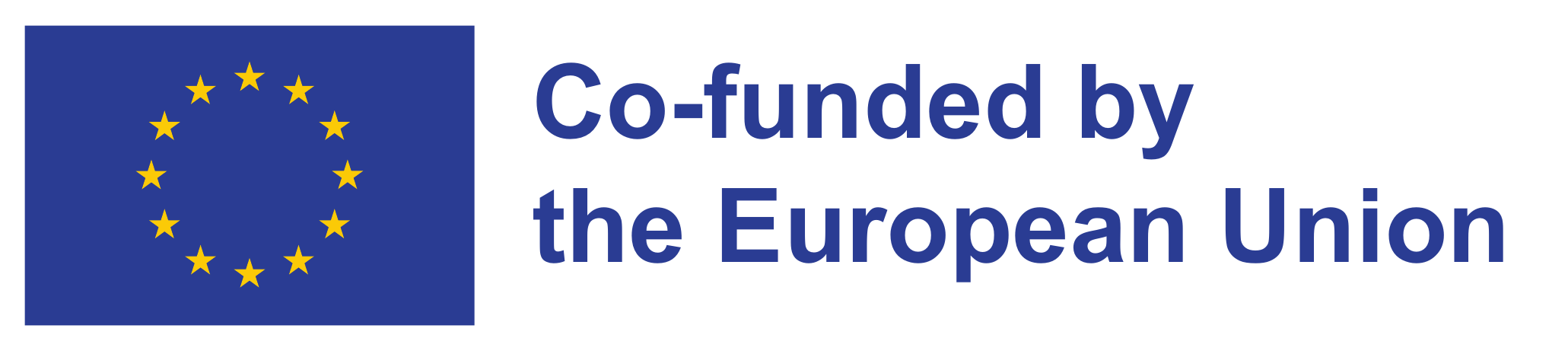}
        \hspace{1em}
    \includegraphics[height=2em]{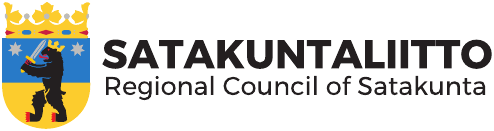}
        \hspace{1em}
    \includegraphics[height=2em]{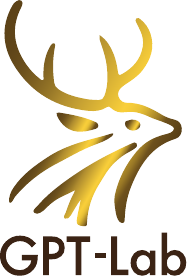}
        \hspace{1em}
    \includegraphics[height=2em]{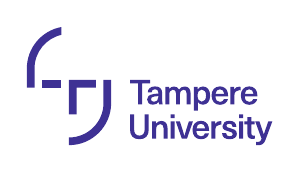}
\end{figure}
%******************************

    \centering
        \vspace{2em}

        \noindent\Large
		\begingroup
			\textbf{\articleTitle}
		\endgroup
  
        \vspace*{0.5em}
        \noindent\large
		\begingroup
			\textbf{\articleSubtitle}
		\endgroup

\vfill

 \noindent\large
            \articleAuthors

\vspace*{0.5em}
 \noindent\normalsize
		\articleAffiliation

        \vspace*{0.5em}
            \articleDate

\vfill

\articleNotes

\vfill

\end{titlepage}

\clearpage

%*******************************************************
% Mainmatter
%*******************************************************
\clearpage
\pagenumbering{arabic}\setcounter{page}{1} % set arabic numbers for main content

%*******************************************************
% Mainmatter
%*******************************************************
% in this example, files are separathed to different "sections"
%\input{your/manuscript/title}
%\input{your/manuscript/introduction}

% you may as well put everything into one single file such as
%!TEX root = ../../main.tex
%********************************
%
\title{TESSA: A Governance-Oriented Custom GPT for SME Guidance under the EU AI Act: Design, Implementation, and Evaluation}

\author{
    Chalisa Veesommai Sillberg\(^a\)
    Kai-Kristian Kemell\(^a\)
    Pekka Sillberg\(^a\)\\
    Katri Harjuveteläinen\(^b\)
    Mika Saari\(^a\)
    Pekka Abrahamsson\(^a\)
    \\
    %\institute{Tampere University, Faculty of Information Technology and Communication Sciences
    \(^a\) Tampere University, Faculty of Information Technology and \\ Communication Sciences,
    Pori, Finland \\
    %\email{\{chalisa.sillberg, kai-kristian.kemell, pekka.sillberg, mika.saari, pekka.abrahamsson\}@tuni.fi}
    \(^b\) Boutique Legal Consulting, Helsinki, Finland  \and 
    \url{chalisa.sillberg,@tuni.fi}
}

\date{\vspace{-5ex}} % removes the automatic date from the document

\maketitle

%\begin{abstract}
%Lorem ipsum dolor sit.
%\keywords{EU AIA, MLR, Ethical AI for Enterprises, Ethics}
%\end{abstract}

\section*{Abstract}
The increasing adoption of generative AI across industries created opportunities for innovation while introducing growing AI governance and regulatory challenges for organization. Small and medium-sized enterprises (SMEs) must navigate compliance requirements under the European Union Artificial Intelligence Act (EU AI Act). Although large language models (LLMs) offer regulatory and educational support, concerns remain regarding governance alignment, reliability, and the practical applicability. To address this challenge, TESSA, a governance-oriented educational artifact, was developed within the OpenAI Custom GPT environment. Using a design science research approach, TESSA was evaluated through a multidimensional framework consisting of governance alignment assessment, human expert review, and consistency testing across 100 benchmark questions in ten thematic categories. The evaluation compared TESSA with a standard ChatGPT baseline using identical benchmark questions and assessment procedures. preliminary LLM-assisted governance alignment reached 90\% for TESSA and 83\% for the baseline, while independent validation by an AI regulation lawyer produced comparable rates of 71\% and 70\%, respectively. Human expert review indicated that both systems provided generally strong governance guidance. Consistency testing showed highly stable performance across repeated executions, with average overall consistency score of 0.969 for TESSA and 0.976 for the baseline. The findings suggest that governance-oriented grounding provides structured and practical compliance guidance while maintaining educational quality and response stability. The study contributes (i) a governance-oriented Custom GPT artifact for AI governance education for SMEs, (ii) a human-validated regulatory update mechanism, and (iii) a multidimensional evaluation framework that can support the evaluation of governance-oriented AI artifact.

\section*{Keywords:} {European Union Artificial Intelligence Act (EU AI Act), AI Governance, Compliance Support, Design Science Research (DSR), Custom GPT}

\section{Introduction}
The rapid adoption of generative AI has accelerated the integration of AI into products, services, and business processes across software businesses and other industries. Consequently, understanding and complying with AI regulations, particularly the EU AI Act, has become an important challenge. This is especially true for small and medium-sized enterprises (SMEs), which often lack dedicated legal expertise and governance resources while pursuing AI-driven innovation. The EU AI Act introduces a risk-based regulatory framework for AI systems and establishes obligations for providers, deployers, importers, distributors, and other stakeholders \cite{Anonymous2026,EUAIAct2024}.

The regulation aims to promote trustworthy AI and protect fundamental rights; however, its complexity creates challenges for SMEs, startups, researchers, and other non-legal stakeholders seeking to understand their responsibilities and compliance obligations.

Existing studies have identified several challenges for SMEs seeking to comply with the EU AI Act, including limited compliance resources, lack of regulatory expertise, uncertainty regarding risk classification and regulatory interpretation, and the need to balance compliance activities with AI system development \cite{duwe2026regulating,kilian2025european,naidu2025designing,stampernas2025framework}. These challenges create a gap between regulatory requirements and practical implementation, highlighting the need for accessible governance support tools. This gap motivated the development of TESSA, a governance-oriented educational artifact designed to translate complex regulatory requirements into practical guidance.

Recent advances in large language models (LLMs) and generative AI have created new opportunities for educational, legal, and compliance support applications. LLMs have increasingly served as the foundation for intelligent support systems across multiple domains. In legal and regulatory contexts, systems such as Chatlaw have demonstrated the potential to provide advisory recommendations and improve access to specialized legal knowledge \cite{cui2026chatlaw}, and interactive legal assistants have been developed to support legal information access and non-specialists seeking legal guidance \cite{yao2025intelligent}. More broadly, AssistantGPT illustrates how LLM-based assistants can bridge the gap between complex system functionalities and intuitive user interaction \cite{neszlenyi2024assistantgpt}, while GPTVoiceTasker represents an example of an LLM-based assistant designed to translate natural language requests into actionable task support \cite{vu2401gptvoicetasker}. However, the reliability of general-purpose LLMs remains a concern in domain-specific applications, motivating the use of domain-specific knowledge integration and grounding mechanisms to improve alignment, reduce hallucinations, and support trustworthy AI-assisted guidance.

Despite growing research on the EU AI Act, AI governance, and LLM-based support systems, limited empirical evidence exists regarding governance-oriented educational artifacts designed to support SME compliance and AI governance learning. To address this gap, we have developed and evaluated TESSA, designed to support EU AI Act education and compliance guidance. The study addresses the following research questions (RQs):

\begin{description}
    \item[RQ1] \textit{To what extent can TESSA support SMEs and other non-legal stakeholders in understanding and applying EU AI Act requirements?} and
    \item[RQ2] \textit{How does governance-oriented grounding influence the governance alignment, educational quality, and consistency of AI-generated compliance guidance compared with baseline ChatGPT?}
\end{description}

This study makes the following contributions: (i) a governance-oriented grounding approach for Custom GPT systems supporting AI governance education, (ii) TESSA, a design science research artifact for SME guidance under the EU AI Act, (iii) a multidimensional evaluation framework, and (iv) an empirical comparison with baseline ChatGPT.

The remainder of the paper is organized as follows. Section \ref{methodology} presents the research methodology, including the TESSA system architecture, governance-oriented grounding strategy, human-validated regulatory update mechanism, and multidimensional evaluation framework. Section \ref{result} reports the evaluation results. Section \ref{discussion} discusses the findings, implications, and limitations of the study. Finally, Section \ref{conclusionandfuturework} concludes the paper and outlines future research directions.

\section{TESSA Design and Research Methodology}
\label{methodology}

\subsection{Research Design}
This study followed the design science research methodology (DSRM) proposed by Peffers et al. \cite{peffers2007design} to design, develop, and evaluate the TESSA artifact. The research problem addressed the challenge faced by SMEs, start-ups, students, researchers, and other non-legal dedicated resources in understanding and operationalizing the requirements of the EU AI Act. Following the DSRM process, the study \textit{identified the practical problem} of SMEs struggling with EU AI Act compliance, \textit{defined the objective} of developing TESSA, \textit{designed and developed} the TESSA artifact within the OpenAI Custom GPT environment, \textit{demonstrated} the artifact by applying it to representative EU AI Act compliance scenarios, and \textit{evaluated} it through governance alignment assessment, human expert review, and consistency evaluation. This paper communicates the initial design and evaluation cycle, providing a foundation for future refinement.

Following the DSR process, the resulting research artifact, TESSA, was designed, implemented, and evaluated to support SMEs and other non-legal stakeholders in interpreting EU AI Act requirements and translating them into practical governance-oriented actions. The evaluation investigated whether governance-oriented grounding improves the alignment, educational usefulness, and consistency of AI-assisted guidance.

\subsection{TESSA System Architecture}

TESSA was developed within OpenAI Custom GPT using the  GPT-5 model available in January 2026. For comparison, the baseline system was the standard ChatGPT interface using the same underlying GPT-5 model without custom instructions, domain-specific knowledge sources, or governance-oriented grounding. TESSA supports users in interpreting EU AI Act requirements and translating them into practical governance-oriented actions. Figure 1 presents the deployed TESSA interface together with its overall system architecture\footnote{The evaluated version of TESSA is available in \textit{link redacted due to double blind review process}}.
The evaluated TESSA configuration consisted of four main components:
\begin{itemize}
    \item System instructions
    \item Published EU AI Act guidance knowledge base \cite{Anonymous2026}
    \item Use case AI List risk classification repository
    \item Human-validated regulatory updates
\end{itemize}

The architecture combines governance-oriented knowledge sources, a structured grounding process, and governance-focused outputs.

\begin{figure}[ht]
	\centering
	\includegraphics[width=0.9\textwidth]{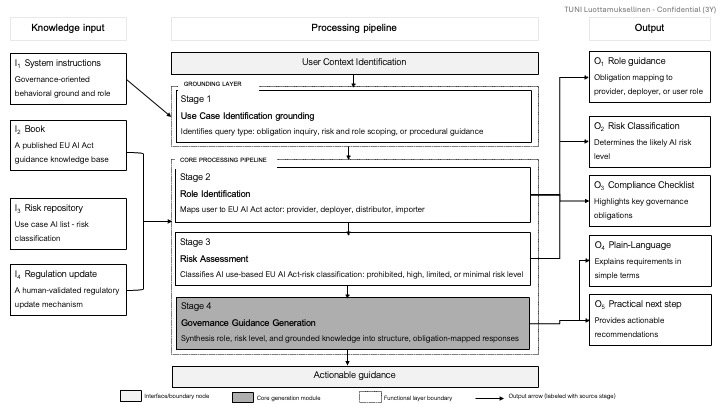}
	\caption{TESSA implementation and system architecture within OpenAI Custom GPT environment.}
	\label{fig:System1}
\end{figure}

\subsection{Governance-Oriented Grounding}

TESSA was designed using a governance-oriented grounding approach to support users in interpreting and applying the requirements of the EU AI Act. Rather than relying solely on general-purpose language model knowledge, the system combines governance-oriented instructions, domain-specific knowledge resources, risk classification support, and human-validated regulatory updates.

When responding to a query, TESSA identifies the relevant use case, stakeholder role, and potential AI risk classification before generating\\ governance-oriented guidance. The process follows the sequence: \textit{Question → Use Case → Role → Risk → Guidance → Output}. The generated output is delivered through the structured governance guidance pattern (SGGP), providing role-specific guidance, risk classification information, compliance considerations, plain-language explanations, and practical recommendations. This grounding approach is intended to improve governance alignment, reduce risk of hallucination, and support actionable compliance guidance.

\subsection{Human-Validated Regulatory Update Mechanism}

As AI regulation and governance guidance continue to evolve, maintaining the currency of regulatory information represents a significant challenge for educational AI artifacts. To maintain current and validated regulatory knowledge, TESSA incorporates a human-validated regulatory update mechanism designed to support the continuous monitoring of relevant regulatory developments while maintaining governance oversight.

The update mechanism consists of two interconnected workflows. The first workflow focuses on automated information collection of candidate regulatory updates, including EU AI governance resources and academic databases, through an N8N-based automation workflow. Collected content is processed, merged, and filtered to remove duplicate entries before being stored in a structured review queue. The second workflow supports human review and controlled publication. Candidate updates are evaluated for relevance, accuracy, and consistency with the governance objectives of the system. Human reviewers assess whether the collected information should be approved or excluded from further use. Only validated updates are subsequently incorporated into TESSA through controlled update processes and associated repositories.

\begin{figure}[ht!]
	\centering
	\includegraphics[width=1\textwidth]{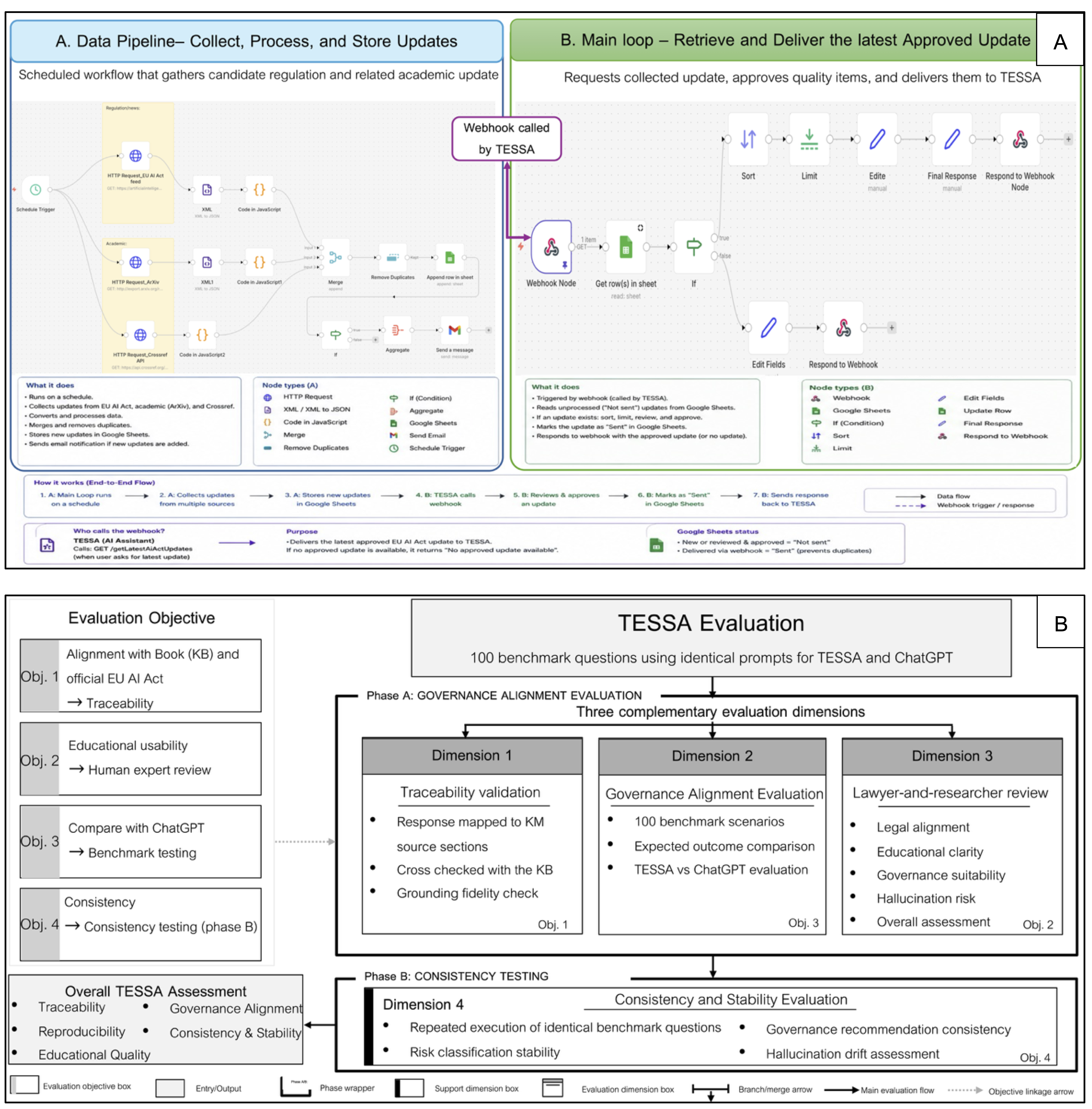}
	\caption{TESSA evaluation and maintenance framework. The figure combines the a) human-validated regulatory update mechanism and b) multidimensional evaluation framework used to assess the governance alignment, educational quality, and response consistency of TESSA across 100 benchmark scenarios.}
	\label{fig:Fig 2 a,b}
\end{figure}

Figure~\ref{fig:Fig 2 a,b} a) represents the overall update mechanism. The process combines automated monitoring with human review to maintain the relevance of the regulatory information used by TESSA. Updates are reviewed before being incorporated into the system, providing an additional quality control step.

\subsection{Multidimensional Evaluation Framework}
The evaluation of TESSA followed a multidimensional framework designed to assess governance alignment, educational suitability, reproducibility, and response consistency. The framework was developed to evaluate i) the correctness of the responses and ii) their usefulness for SME-oriented compliance education and governance support. The evaluation consisted of four primary objectives (Obj):
\begin{enumerate}
    \item To assess whether TESSA responses aligned with the published EU AI Act guidance knowledge base \cite{Anonymous2026} and official EU AI Act sources \cite{EUAIAct2024}.
%    Navigating Compliance: knowledge base and official EU AI Act sources.
    \item To evaluate the educational suitability and governance quality of the responses.
    \item To compare TESSA with baseline ChatGPT using identical benchmark scenarios.
    \item To assess response consistency across repeated executions of the same scenarios.
\end{enumerate}

To address these objectives, a multidimensional evaluation framework comprising four complementary dimensions was adopted: traceability validation, governance alignment evaluation, human expert review, and consistency evaluation. As shown in Figure~\ref{fig:Fig 2 a,b} b) the framework consists of two evaluation phases. Phase A focuses on governance alignment through i) traceability validation, ii) governance alignment evaluation, and iii) human expert review. Phase B evaluates response consistency through repeated execution of benchmark scenarios.

The evaluation dimensions were informed by design science research (DSR) evaluation principles \cite{venable2016feds}, trustworthy AI concepts reflected in the NIST AI Risk Management Framework \cite{ai2023artificial}, and structured qualitative assessment approaches. Governance alignment, legal alignment, educational clarity, governance suitability, and hallucination risk were assessed using a three-level rating structure adapted from QualSyst \cite{kmet2004standard}.

\subsubsection{Benchmark Design and Traceability Validation}% was 3.5.1
The benchmark scenarios were generated through a GPT-assisted process and subsequently reviewed and curated by the first author. The objective was to create realistic educational, governance, and compliance-related questions that SMEs, startups, students, researchers, and other non-legal stakeholders might ask when seeking guidance on the EU AI Act. 

The benchmark was organized into ten categories: (i) AI literacy, (ii) startup and student projects, (iii) risk classification, (iv) compliance duties, (v) prohibited AI practices, (vi) high-risk AI, (vii) limited-risk AI, (viii) minimal-risk AI, (ix) GPAI and foundation models, and (x) SME perspectives, each containing ten questions, resulting in a total of 100 evaluation scenarios. Following scenario generation, responses were produced using TESSA and baseline ChatGPT. The responses were then evaluated against expected outcomes developed from a published EU AI Act guidance knowledge base \cite{Anonymous2026} and relevant official EU AI Act sources \cite{EUAIAct2024}. This traceability validation supported Objective 1 by assessing the alignment between responses and the grounding sources.

\subsubsection{Governance Alignment Evaluation}%was 3.5.2

A governance alignment evaluation was conducted to assess responses aligned with expected outcomes derived from a published EU AI Act guidance knowledge base \cite{Anonymous2026} and official EU AI Act sources \cite{EUAIAct2024}, and served as the benchmark for governance alignment assessment. This evaluation primarily addressed Obj. 1 and the supporting Obj. 3 by comparing TESSA and baseline ChatGPT under identical benchmark conditions.

Responses from 100 benchmark scenarios were compared with the corresponding expected outcome. The evaluation considered governance interpretation, risk classification, compliance recommendations, and consistency with the reference sources.
Responses were assessed using the three-level alignment scale (Yes, Partial, No) presented in Table~\ref{table:EvaluationCriteria}. The scoring approach was adapted from the QualSyst framework \cite{kmet2004standard}. 

The evaluation consisted of two stages. First, an LLM-assisted preliminary review supported the initial classification of the responses. Second, the classifications were independently reviewed and validated by an experienced AI regulation lawyer practitioner with more than five years of professional legal experience in AI and data protection. The lawyer-reviewed assessments were used as the final governance alignment results reported in this study.
The evaluation was intended to assess alignment with predefined educational and governance expectations rather than establish definitive legal interpretations. The resulting assessments enabled systematic comparison between TESSA and baseline ChatGPT and formed one component of the multidimensional evaluation framework.

\begin{table}[ht!]
    \caption{Multidimensional Evaluation Criteria}
    \centering
    \scriptsize%added on 2024-08-22
    \begin{tabularx}{\textwidth}{ p{1.125cm} p{1.25cm} X}
   \toprule
  Category & Criteria  			& Description\\%[0.5ex]
  \midrule

\multicolumn{3}{l}{\textbf{1. Ground truth}} \\
             &Yes & The response aligned with the expected outcome. \\
 			&Partial & The response was generally aligned but incomplete, lacked important nuance, or omitted relevant governance considerations.\\
 			& No & The response was incorrect, misleading, or substantially inconsistent with the expected outcome.\\ \midrule

\multicolumn{3}{l}{\textbf{2. Human expert review}} \\
\multicolumn{3}{l}{2.1~Legal alignment} \\
                                         &Strong & Legally consistent and aligned with EU AI Act concepts and official sources.\\
                                         &Acceptable & Mostly aligned but contains minor ambiguity or incomplete interpretation.\\
                                         &Weak & Legally misleading, incorrect, or significantly inconsistent with official sources.\\
\multicolumn{3}{l}{2.2~Educational clarity} \\
                                         &Strong & Clear, accessible, and understandable for SMEs and non-legal users.\\
                                         &Acceptable & Generally understandable but may contain unclear or overly complex parts.\\
                                       & Weak & Confusing, poorly structured, or difficult for intended users to understand.\\
\multicolumn{3}{l}{2.3~Governance suitability} \\
                                        &Strong & Promotes resonsible and proportionate governance practices.\\
                                        & Acceptable & Generally appropriate but lacks nuance or practical guidance.\\
                                        & Weak & Inappropriate, insufficient, or potentially irresponsible governance guidance.\\
\multicolumn{3}{l}{2.4~Hallucination risk} \\
                                        & Strong & No significant unsupported or fabricated claims detected.\\
                                        & Acceptable & Minor unsupported assumptions or over-generalizations present.\\
                                       & Weak & Contains fabricated, misleading, or unsupported legal/compliance claims.\\
\multicolumn{3}{l}{2.5~Overall Assessment} \\
                                        & Strong & Reliable and appropriate for educational compliance guidance.\\
                                        & Acceptable & Mostly reliable with some limitations.\\
                                        & Weak & Insufficiently reliable or potentially misleading. \\ \midrule
\multicolumn{3}{l}{\textbf{3. Consistency Evaluation}} \\
\multicolumn{3}{l}{3.1~Consistency}\\
            & Stable & Semantically consistent; no meaningful legal/governance variation. \\
 			& Minor variation & Small detail differences; overall governance/legal meaning aligned. \\
 			& Major variation & Contradictory classification, divergent governance, or scenario mismatch. \\
\multicolumn{3}{l}{3.2~Hallucination drift} \\
            & Low & Semantically consistent; no meaningful legal/governance variation. \\
 			& Medium & Occasional unsupported or exaggerated claims.\\
 			& High & Repeated unsupported claims, scenario misidentification, or contradictory reasoning.\\
            \midrule
    \end{tabularx}
    \label {table:EvaluationCriteria}
    \end{table}

\subsubsection{Human Expert Review}
In addition to benchmark-based evaluation, a human expert review was conducted to assess the educational and governance quality, as well as the legal interpretation of the responses. This evaluation supported Obj. 2 of the evaluation framework by examining whether the responses were suitable for the intended users seeking guidance on the EU AI Act.
The review was conducted using a structured assessment framework consisting of five dimensions: (i) legal alignment, (ii) educational clarity, (iii) governance suitability, (iv) hallucination risk, and (v) overall assessment. The detailed assessment criteria are presented in Table~\ref{table:EvaluationCriteria}. Similar to the governance alignment evaluation, qualitative expert assessments were recorded using a three-level rating structure consisting of strong, acceptable, and weak. The framework was designed to assess the factual and regulatory consistency of the responses as well as their usefulness for compliance education and governance support.

The review process consisted of two stages. First, an LLM-assisted preliminary assessment was reviewed by the first author, a postdoctoral researcher with two years of research experience in AI governance and the EU AI Act. Second, legal interpretation and governance-related assessments were independently validated by an AI regulation law practitioner with more than five years of professional legal experience in AI and data protection. 

\subsubsection{Consistency Evaluation} 

Consistency is an important quality attribute for evaluating design artifacts \cite{venable2016feds}. To assess reproducibility, each of the 100 benchmark scenarios was executed three times using both TESSA and baseline ChatGPT under identical prompting conditions, resulting a total of 600 responses.

Five dimensions were evaluated: (i) risk classification consistency, ( ii) governance recommendation consistency, (iii) legal interpretation consistency, (iv) hallucination drift, and (v) reasoning variability. Responses were assessed using a three-level scale consisting of stable, minor variation, and major variation. Ratings were converted into numerical values (2–1–0) following adaptation of the QualSyst scoring logic \cite{kmet2004standard} and aggregated into a normalized overall consistency score (OCS):

\begin{equation}
OCS = \frac{\text{total score obtained}}{\text{maximum possible score}}
\end{equation}

Higher OCS values indicate greater consistency across repeated executions. Detailed assessment criteria are presented in Table~\ref{table:EvaluationCriteria}. Benchmark prompts, responses, reviewer assessments, and consistency records were documented to support traceability and reproducibility.

\subsubsection{Data Collection and Analysis} % was 3.5.6

Data collected across both evaluation phases included response summaries, reviewer assessments, and consistency ratings, which were analyzed using quantitative and qualitative methods to compare TESSA and baseline ChatGPT.

\section{Results}
\label{result}

This section presents the empirical evaluation results obtained using the multidimensional evaluation framework described in Section 2.5. The design and implementation of TESSA artifact were presented in the previous section: therefore, the following results focus on evaluating governance alignment, human expert review, and consistency. The results compare TESSA with baseline ChatGPT using identical benchmark scenarios and evaluation procedures.

\subsection{Governance Alignment and Traceability Results} 

The governance alignment evaluation assessed the extent to which responses aligned with predefined expected outcomes derived from i) a published EU AI Act guidance knowledge base \cite{Anonymous2026}
and ii) official EU AI Act sources \cite{EUAIAct2024}. A total of 100 benchmark scenarios covering ten thematic categories were evaluated using the "Yes", "Partial", and "No" alignment framework.

The reported results reflect the final lawyer-validated assessment. An LLM-assisted preliminary review was initially conducted to support classification, producing alignment rates of 90\% for TESSA and 87\% for baseline ChatGPT. The responses were subsequently reviewed and validated by an experienced AI regulation lawyer, and the validated classifications are reported in Table~\ref{tab:alignment_results}.

\begin{table}[ht!]
\centering
\caption{Governance Alignment Results}
\label{tab:alignment_results}
\begin{tabular}{lcccc}
\toprule
System & Yes & Partial & No & Full Alignment (\%) \\
\midrule
LLM-assisted preliminary review & & & \\
-- TESSA & 90 & 10 & 0 & 90 \\
-- Baseline ChatGPT & 83 & 17 & 0 & 83 \\
\midrule

Experienced AI regulation lawyer & & & \\
-- TESSA & 71 & 27 & 2 & 71 \\
-- Baseline ChatGPT & 70 & 30 & 0 & 70 \\
\bottomrule
\end{tabular}
\end{table}

TESSA achieved an overall alignment rate of 71\%, compared with 70\% for baseline ChatGPT. Most differences occurred in scenarios where responses were generally correct but lacked governance-related nuance, stakeholder-specific obligations, or practical compliance considerations. The lawyer review applied a stricter interpretation of governance expectations than the preliminary LLM-assisted assessment, resulting in several responses being reclassified from full alignment to partial alignment.

The findings indicate that both systems were capable of generating educational guidance related to the EU AI Act. However, TESSA demonstrated slightly stronger alignment with predefined governance expectations and source-based interpretations. The governance-oriented grounding approach appeared particularly beneficial in scenarios involving stakeholder responsibilities, risk classification, and practical compliance actions.

From a traceability perspective, responses generated by TESSA more consistently reflected concepts and terminology contained within the grounding sources. This supports Objective 1 of the evaluation framework, which sought to assess alignment with the published EU AI Act guidance knowledge base \cite{Anonymous2026} and official EU AI Act sources \cite{EUAIAct2024}.

\subsection{Human Expert Review Results}

The human expert review evaluated responses generated by TESSA and baseline ChatGPT using the framework described in Section 2.5.3. The final results reported in this section are based on the review conducted by an experienced AI regulation lawyer, while a separate AI regulation researcher review was used to support the assessment process.

Table~\ref{tab:human_review_results} summarizes the lawyer-reviewed results. Both TESSA and the baseline ChatGPT system demonstrated strong performance across the evaluation dimensions. Educational clarity was the strongest category for both systems. TESSA received 99 "Strong" ratings and one "Acceptable" rating, while baseline ChatGPT received 92 "Strong" ratings and eight "Acceptable" ratings. No responses were classified as "Weak" in this category, indicating that both systems consistently produced explanations that were understandable and accessible for the intended users.

The review also identified differences in emphasis between the two systems. Baseline ChatGPT received higher "Strong" ratings in Legal Alignment and Overall Assessment, whereas TESSA received higher "Strong" ratings in Governance Suitability.

\begin{table*}[t]
\centering
\caption{Human Expert Review Results}
\label{tab:human_review_results}
\begin{tabular}{lcccccc}
\toprule
\multirow{2}{*}{Criterion}
& \multicolumn{3}{c}{TESSA}
& \multicolumn{3}{c}{Baseline ChatGPT} \\
\cline{2-7}
& Strong & Acceptable & Weak
& Strong & Acceptable & Weak \\
\midrule
\multicolumn{7}{l}{\textbf{Experienced AI regulation lawyer}} \\
-- Legal Alignment & 53 & 44 & 3 & 67 & 32 & 1 \\
-- Educational Clarity & 99 & 1 & 0 & 92 & 8 & 0 \\
-- Governance Suitability & 84 & 13 & 3 & 76 & 23 & 1 \\
-- Hallucination Risk Assessment & 61 & 36 & 3 & 78 & 21 & 1 \\
(low, medium, high) &  &  &  &  &  &  \\
Overall Assessment & 59 & 37 & 4 & 75 & 24 & 1 \\
\midrule

\multicolumn{7}{l}{\textbf{AI regulation researcher}} \\
-- Legal Alignment & 70 & 30 & 0 & 83 & 17 & 0 \\
-- Educational Clarity & 99 & 1 & 0 & 99 & 1 & 0 \\
-- Governance Suitability & 87 & 13 & 0 & 83 & 17 & 0 \\
-- Hallucination Risk Assessment & 71 & 29 & 0 & 87 & 13 & 0 \\
(low, medium, high) &  &  &  &  &  &  \\
Overall Assessment & 70 & 30 & 0 & 83 & 17 & 0 \\
\bottomrule
\end{tabular}
\end{table*}

Hallucination risk was generally low for both systems. TESSA received 61 low-, 36 medium-, and 3 high-risk ratings, while the baseline ChatGPT system received 78 low-, 21 medium-, and 1 high-risk ratings. The higher proportion of low risk ratings for baseline ChatGPT indicates fewer unsupported assumptions or over-generalized statements. Across both systems, medium- and high-risk assessments primarily reflected unsupported assumptions, incomplete explanations, or generalized statements rather than fabricated legal obligations or substantially incorrect regulatory interpretations.

Similarly, the first author's review found qualitative differences in emphasis between the two systems. Overall, the human expert review indicates that the two systems exhibited different strengths rather than one consistently outperforming the other. Baseline ChatGPT received higher ratings in Legal Alignment, Hallucination Risk, and Overall Assessment, whereas TESSA received higher Governance Suitability ratings, reflecting its stronger emphasis on stakeholder responsibilities and practical compliance guidance.

This suggests that governance-oriented grounding influenced the focus of TESSA responses toward stakeholder responsibilities, governance considerations, and compliance-oriented recommendations.

\subsection{Consistency Results}

The consistency evaluation assessed whether repeated executions of identical benchmark scenarios produced stable governance reasoning, risk classifications, and compliance recommendations. Each of the 100 benchmark scenarios was executed three times using both TESSA and baseline ChatGPT, resulting in a total of 600 responses.
Table \ref{tab:stability_results} summarizes the consistency evaluation results. Both systems demonstrated a high degree of consistency across repeated executions. TESSA achieved an average OCS of 0.969, while baseline ChatGPT achieved an average OCS of 0.976. The difference between the systems was negligible ($\Delta OCS = -0.007$), indicating broadly comparable consistency performance across the benchmark dataset.

\begin{table}[ht]
\centering
\caption{Consistency Results}
\label{tab:stability_results}
\begin{tabular}{lcc}
\toprule
Metric & TESSA & Baseline ChatGPT \\
\midrule
Stable & 98 & 98 \\
Minor Variation & 0 & 1\\
Major Variation & 2 & 1 \\
Average OCS & 0.969 & 0.976 \\
Standard Deviation OCS & 0.139 & 0.108 \\
OCS $\geq$ 0.75(\%) & 98 & 99\\
OCS $\geq$ 0.90(\%) & 94 & 96 \\
$\Delta OCS$ (TESSA-Baseline ChatGPT) & -0.007 & \\
Low Hallucination Drift (\%) & 98\% & 99\% \\
\bottomrule
\end{tabular}
\end{table}

For TESSA, 98\% of the scenarios were classified as "Stable" and two scenarios as "Major Variation". For baseline ChatGPT, 98\% of scenarios were classified as "Stable", one scenario as "Minor Variation", and one scenario as "Major Variation". Most variations involved differences in explanation style, emphasis, or recommendation detail rather than substantial changes in legal interpretation, governance guidance, or risk classification.
Hallucination drift remained low for both systems. TESSA achieved a low hallucination drift rate of 98\%, while baseline ChatGPT achieved 99\%. No evidence of systematic hallucination drift was observed across repeated executions. The few identified variations primarily reflected minor unsupported assumptions or differences in explanatory detail rather than fabricated compliance obligations or contradictory governance recommendations.

Overall, the results indicate that both TESSA and baseline ChatGPT maintained highly consistent performance across repeated executions. The findings further suggest that the governance-oriented grounding approach implemented within TESSA did not negatively affect response consistency and was able to maintain stable governance reasoning across multiple runs.

\subsection{Summary of Findings}

The evaluation demonstrated that both TESSA and baseline ChatGPT were capable of providing educational guidance related to the EU AI Act. Governance alignment was comparable between the two systems. TESSA received higher Governance Suitability ratings, while baseline ChatGPT received higher Legal Alignment and Overall Assessment ratings.

Educational Clarity was a particular strength of both systems, indicating that EU AI Act concepts could be communicated effectively to the intended users.

Both systems also demonstrated highly consistent performance across repeated executions, achieving comparable OCS values and low hallucination drift. The findings suggest that governance-oriented grounding provided structured governance guidance without reducing response consistency. 
Taken together, the results provide preliminary answers to the research questions. For RQ1, TESSA was able to translate complex EU AI Act requirements into understandable and actionable guidance. For RQ2, governance-oriented grounding contributed to structured governance-oriented educational guidance while maintaining consistency and low hallucination risk.

\section{Discussion}
\label{discussion}

\subsection{From Regulatory Complexity to Actionable Guidance}

One objective of TESSA was to help SMEs and other non-legal stakeholders interpret EU AI Act requirements and translate them into practical governance actions. The evaluation results suggest that this objective was achieved. High ratings for Educational Clarity and Governance Suitability indicate that TESSA was able to provide understandable explanations while supporting practical governance-oriented decision making. The governance-oriented grounding approach, which combines role identification, risk assessment, and domain-specific knowledge sources, appears to support the transformation of complex regulatory requirements into actionable guidance for resource-constrained stakeholders.

\subsection{Governance-Oriented Grounding Compared with Baseline ChatGPT}

The comparison with baseline ChatGPT revealed different strengths between the two systems. The comparable legal performance observed in this study may reflect the increasing availability of EU AI Act knowledge in modern foundation models and other publicly available regulatory sources. Although governance alignment was comparable, between the two systems TESSA received higher Governance Suitability ratings, suggesting that\\ governance-oriented grounding improved alignment with compliance-oriented educational objectives. 

In contrast, baseline ChatGPT often received higher Legal Alignment and Overall Assessment ratings, reflecting its ability to provide fluent and broadly applicable explanations. These findings suggest that the primary contribution of governance-oriented grounding is the structured presentation of governance guidance rather than improving general legal knowledge.
Baseline ChatGPT also performed well, suggesting that the primary contribution of TESSA lies in its structured governance guidance rather than exclusive access to regulatory knowledge.

\subsection{Implications for SMEs and Non-Legal Stakeholders}

The findings have practical implications for SMEs, students, researchers, and other non-legal stakeholders affected by the EU AI Act. TESSA may help bridge the gap between regulatory requirements and practical implementation by providing structured explanations, risk-classification support, and stakeholder-specific guidance. Rather than replacing legal professionals, systems such as TESSA can function as educational support tools that improve understanding of governance responsibilities, potential risks, and compliance considerations.

\subsection{Trustworthy AI and Human-Validated Regulatory Updates}

A key feature of TESSA is the incorporation of a human-validated regulatory update mechanism. By combining automated monitoring of regulatory developments with human review, the mechanism supports the integration of updated information while maintaining human oversight. This approach aligns with trustworthy AI principles and contributes to the reliability of governance-oriented educational guidance.

\subsection{Limitations}

Several limitations should be considered when interpreting the findings of this study.
First, TESSA and the baseline GPT were evaluated during the same evaluation period (April-June, 2026) using the latest and same underlying model version available at the time of testing. Future model updates may influence performance and affect comparative results.
Second, the Custom GPT platform did not provide direct API access for automated benchmarking. Consequently, responses were collected and evaluated manually, increasing evaluation effort and introducing a potential risk of recording errors.
Third, the evaluation involved human judgment when assessing governance alignment and response quality. Part of the human expert review was conducted by the first author, introducing a potential source of evaluator bias. To mitigate this risk, a structured evaluation framework, an LLM-assisted preliminary review, and independent legal validation were employed. Nevertheless, some degree of subjectivity remains unavoidable.
Finally, the study evaluated a single governance-oriented Custom GPT configuration focused on EU AI Act education and compliance support. Additional studies involving other regulatory frameworks, deployment contexts, and user populations would strengthen the generalizability of the findings.

\section{Conclusion and Future Work}
\label{conclusionandfuturework}

In this study, we developed and evaluated TESSA to support SMEs and other non-legal stakeholders in understanding and applying the EU AI Act. Using a multidimensional evaluation framework, we compared TESSA with baseline ChatGPT across 100 benchmark scenarios. The results indicated that TESSA provided strong governance-oriented guidance, high educational clarity, and stable performance across repeated executions. The findings suggest that governance-oriented grounding can improve the practical applicability of AI-generated compliance guidance while maintaining response consistency.

Future work will focus on extending TESSA beyond the EU AI Act by integrating additional AI regulations and governance frameworks into a unified knowledge environment and evaluating its use in real-world settings.

\subsubsection*{Acknowledgments}
This work was funded by the European Regional Development Fund and the Regional Council of Satakunta and Business Finland through the Co-Innovation project AI Champion.

\subsubsection*{Declaration of competing interest}
The authors have no competing interests to declare that are relevant to the content of this article.

\subsubsection*{Data availability}

The supplementary dataset used to evaluate TESSA is available on Zenodo at \url{https://doi.org/10.5281/zenodo.22874463}. It contains 100 benchmark questions across ten thematic evaluation categories, expected outcomes, TESSA and baseline ChatGPT responses, and item-level preliminary and lawyer-validated classifications.

\subsubsection*{AI Use Statement} AI tools (Quillbot/ChatGPT) were used only for grammar editing. The authors remain fully responsible for the scientific content, methodology, analysis, interpretation, and final manuscript.
% setup your included files here

% ******************************************************
% Backmatter
%*******************************************************

%%% Bibliography section
% BibLaTeX: This is accompanying "Option A" on printing the bib items with BibLaTeX
\printbibliography\addcontentsline{toc}{chapter}{References}

% Bibtex/Natbib: This is "Option B" of adding and printing references with BibTex
%\bibliographystyle{plain}
%\bibliography{main.bib}

\end{document}